\documentclass[sigconf]{acmart}

\copyrightyear{2026}
\acmYear{2026}
\setcopyright{cc}
\setcctype{by}
\acmConference[WWW '26] {Proceedings of the ACM Web Conference 2026}{April 13--17, 2026}{Dubai, United Arab Emirates.}
\acmBooktitle{Proceedings of the ACM Web Conference 2026 (WWW '26), April 13--17, 2026, Dubai, United Arab Emirates}
\acmISBN{979-8-4007-2307-0/2026/04}
\acmDOI{10.1145/3774904.3792240}

\AtBeginDocument{%
  }

\usepackage[utf8]{inputenc}
\DeclareUnicodeCharacter{2212}{\ensuremath{-}}
\usepackage{courier}  
\usepackage{natbib}  
\usepackage{caption} 
\usepackage{svg} 
\usepackage[table,dvipsnames]{xcolor}
\usepackage{algorithm}
\usepackage{algorithmic}
\usepackage{multirow}
\usepackage{listings}
\usepackage{makecell}
\usepackage{booktabs}
\usepackage{enumitem}
\usepackage{amsmath}
\usepackage{amsfonts}
\usepackage{cleveref}
\usepackage{listings}      
\usepackage{xcolor}        

\usepackage{tcolorbox}     
\tcbuselibrary{listings}   
\tcbuselibrary{breakable}  

\begin{document}

\title{Difficulty-Aware Agentic Orchestration for Query-Specific Multi-Agent Workflows}


\author{Jinwei Su}
\authornote{Equal contribution.}
\orcid{0009-0001-9208-3354}
\affiliation{
  \institution{South China Normal University}
  \city{Guangzhou}
  \country{China}
}
\email{2024024579@m.scnu.edu.cn}

\author{Qizhen Lan}
\authornotemark[1]
\orcid{0000-0003-0496-5240}
\affiliation{
  \institution{The University of Texas Health Science Center}
  \city{Houston}
  \country{United States}
}
\email{Qizhen.Lan@uth.tmc.edu}

\author{Yinghui Xia}
\authornote{Corresponding authors.}
\orcid{0009-0000-4494-6074}
\affiliation{
  \institution{The Hong Kong University of Science and Technology}
  \city{Guangzhou}
  \country{China}
}
\email{yxia501@connect.hkust-gz.edu.cn}

\author{Lifan Sun}
\orcid{0009-0007-3126-7196}
\affiliation{
  \institution{University of California}
   \city{San Diego}
  \country{United States}
}
\email{lis005@ucsd.edu}

\author{Weiyou Tian}
\orcid{0009-0004-4397-3897}
\affiliation{
  \institution{Peking University}
  \city{Shenzhen}
  \country{China}
}
\email{weiyoutian25@stu.pku.edu.cn}

\author{Tianyu Shi}
\authornotemark[2]
\orcid{0009-0001-9119-778X}
\affiliation{
  \institution{University of Toronto}
  \city{Toronto}
  \country{Canada}
}
\email{tys@cs.toronto.edu}

\author{Lewei He}
\authornotemark[2]
\orcid{0009-0003-0510-911X}
\affiliation{
  \institution{South China Normal University}
  \city{Guangzhou}
  \country{China}
}
\email{helewei@m.scnu.edu.cn}

\author{Yang Jingsong}
\authornotemark[2]
\affiliation{
  \institution{Autoagent.AI}
  \city{Beijing}
  \country{China}
}
\email{edward.yang@agentspro.cn}


\renewcommand{\shortauthors}{Jinwei Su et al.}

\begin{abstract}
Large Language Model (LLM)-based agentic systems have shown strong capabilities across various tasks. However, existing multi-agent frameworks often rely on static or task-level workflows, which either over-process simple queries or underperform on complex ones, while also neglecting the efficiency-performance trade-offs across heterogeneous LLMs. To address these limitations, we propose \textbf{Difficulty-Aware Agentic Orchestration (DAAO)}, which can dynamically generate query-specific multi-agent workflows guided by predicted query difficulty. 
DAAO comprises three interdependent modules: a variational autoencoder (VAE) for difficulty estimation, a modular operator allocator, and a cost- and performance-aware LLM router. 
A self-adjusting policy updates difficulty estimates based on workflow success, enabling simpler workflows for easy queries and more complex strategies for harder ones.
Experiments on six benchmarks demonstrate that DAAO surpasses prior multi-agent systems in both accuracy and inference efficiency, validating its effectiveness for adaptive, difficulty-aware reasoning.Our code is open-sourced at \textit{https://github.com/AutoAgents-ai/DAAO} 
\end{abstract}




\maketitle

\begin{figure*}[!ht]
\centering
\includegraphics[width=1\linewidth]{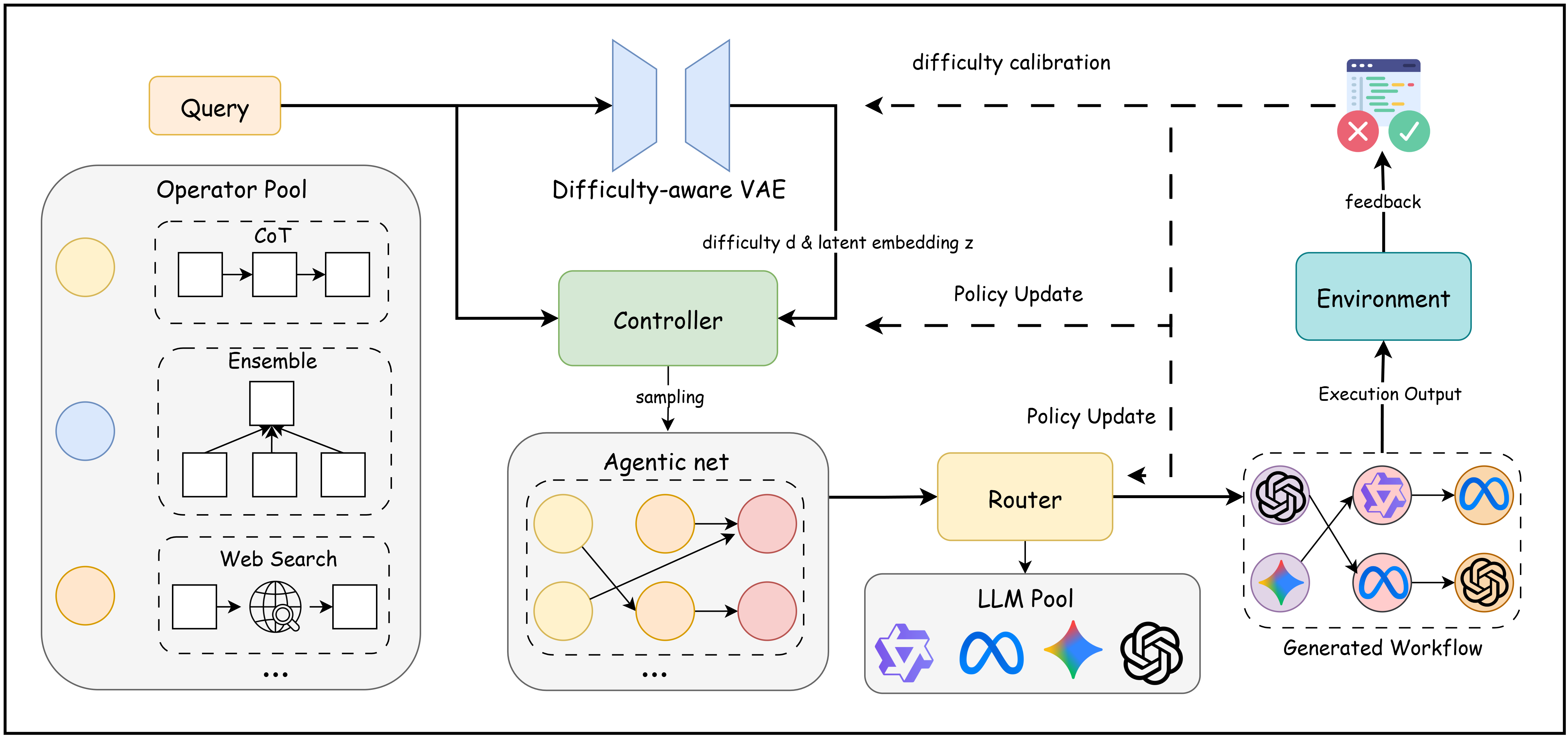}
\caption{The overall framework of our proposed DAAO.}
\label{fig:framework}
\end{figure*}

\section{Introduction}
Large Language Model (LLM)-based agents~\citep{autogpt,babyagi,agentgpt,lou2025drf} have exhibited remarkable capabilities across a wide spectrum of tasks, including question answering~\cite{zhu2024autotqa}, data analysis~\citep{hong2024datainterpreter,li2024autokaggle,shen2025ai}, decision-making~\citep{song2023llmplanner,liu2026health,sun2025objective}, code generation~\citep{reflexion} and web navigation~\citep{deng2024mind2web}. Building upon the success of single agents, recent advancements reveal that organizing multiple LLM-based agents into structured agentic workflows can significantly enhance task performance. In such workflows, agents can interact either cooperatively~\cite{zhuge2024gptswarm} or competitively~\cite{zhao2023competeai} depending on the task context. These multi-agent systems can overcome the cognitive and functional limitations of individual models~\citep{arXiv2023_MultiAgent-Debate,arXiv2023_MultiAgent-Debate_2,multi-persona,blender,autogen,zhang2024cut}, thereby exhibiting collective intelligence similar to human collaboration in a society of agents.

In recent years, the research community has focused on automating multi-agent system design. For instance, DsPy~\citep{khattab2023dspy} and EvoPrompting~\citep{guo2023evoprompt} automate prompt optimization, GPTSwarm~\citep{zhuge2024gptswarm} optimizing inter-agent communication,
and EvoAgent~\citep{yuan2024evoagent} self-evolving agent profiling. However, these systems are often constrained by limited search spaces and rigid representation paradigms, resulting in marginal performance gains and limited adaptability to diverse task requirements. Subsequently, ADAS~\cite{hu2024adas} and AFlow~\citep{zhang2024aflow} employ code as representation for workflow, facilitating robust and flexible workflow searches through different paradigms, with ADAS utilizing heuristic search and AFlow adopting Monte Carlo tree search. MaAS~\citep{Zhang2025MultiagentAS} proposes an agentic supernet to generate a query-specific multi-agent system for each user query. 

How to dynamically generate a workflow given a query remains a key challenge in current research. Task-level workflows~\cite{hu2024adas, zhang2024aflow} are typically built as uniform multi-agent systems for entire task categories, achieving strong metrics like accuracy and pass@k but relying on heavy pipelines with excessive LLM calls and tool usage. This design over-processes simple queries, wasting resources and overlooking factors like token cost and latency. Query-level workflows~\citep{Zhang2025MultiagentAS} introduce input-specific adaptation, but their granularity is often insufficient, leading to suboptimal or oversimplified workflows for difficult inputs. 
For instance, when a user requests a travel guide for a specific location, a workflow that only retrieves and summarizes information often falls short of meeting the user’s needs.
These limitations motivate a difficulty-adaptive framework that dynamically balances complexity and cost. 


To address the above challenges, we propose \textbf{Difficulty-Aware Agentic Orchestration (DAAO)}, which can intelligently generate workflows according to the characteristics of each query. DAAO has three core capabilities:
\textbf{(1)} Learning to capture the difficulty of each query from posterior knowledge, without relying on manual labels;
\textbf{(2)} Dynamically creating workflows that match the predicted difficulty of each query;
\textbf{(3)} Assigning LLMs to workflow components to maximize the reasoning ability of the Multi-agent system.

Technically, we define query difficulty as a learnable policy. Unlike previous methods, LLMs have little knowledge about workflow generation, and manually creating query-workflow pairs requires much human effort, which goes against automatic workflow generation. To address this, We use a reward-like mechanism to update the policy. When a workflow successfully solves a query, we slightly lower its predicted difficulty, allowing future workflows to be simpler. If a workflow fails, we increase the predicted difficulty to encourage more complex and capable workflows. 
In addition, we model multi-agent workflow generation on the agentic net, a probabilistic, continuous agentic architecture distribution that encompasses a vast number of possible multi-agent candidates. To enhance efficiency and adaptability, we incorporate a cost- and performance-aware LLM router that dynamically assigns heterogeneous models to different operators according to query difficulty and resource constraints.
During training, a controller network samples multi-agent architectures conditioned on the input query and updates its policy through feedback signals.
During inference, for different queries, DAAO samples a suitable multi-agent system delivering satisfactory resolution and appropriate inference resources. 

Our key contributions are as follows:
\begin{itemize}[noitemsep, topsep=0pt]
    \item \textbf{Query-Level Difficulty Estimation for Workflow Adaptation.} We propose generating adaptive workflow strategies guided by query difficulty. The framework learns to represent the latent difficulty space of queries, leveraging workflow feedback to adjust the strategies.
    \item \textbf{Dynamic Workflow Generation.} We propose DAAO, a difficulty-aware framework that dynamically generates workflows and realizes LLM heterogeneity based on query difficulty, domain, and features, while achieving performance-cost balance.
    \item \textbf{Experimental Validation.} We conduct comprehensive evaluations on six widely adopted benchmarks, covering diverse use cases in code generation (HumanEval, MBPP), mathematical reasoning (GSM8K, MATH), knowledge and reasoning understanding (MMLU) and diverse tool usage (GAIA). Empirical results demonstrate that DAAO is \textbf{ (1) highly performing}, surpassing existing automated orchestration methods by $3.5\% \sim 15.2\%$ and recent LLM routing methods by $3.2\% \sim 10.2\%$; \textbf{(2) economical}, outperforming the SOTA baseline MasRouter on the MATH benchmark with 65\% of the training cost and 41\% of the inference cost; \textbf{(3) inductive}, demonstrating strong generalization to unseen LLM backbones and transferability across diverse datasets.
\end{itemize}

\section{Related Work}
\paragraph{Automated Agentic Workflows.}  
The development of agentic workflows has evolved from manual configurations to automated systems, with the latter offering improved adaptability and task performance. Early approaches to automation focus on optimizing prompt structures and inter-agent communication protocols~\citep{khattab2023dspy, guo2023evoprompt, zhuge2024gptswarm, yuan2024evoagent}, thereby enhancing the robustness of workflows across a variety of tasks. More recent systems, such as ADAS~\citep{hu2024adas} and AFlow~\citep{zhang2024aflow}, leverage code-based representations to enable real-time structural adaptation and communication strategy refinement based on environmental feedback. MaAS~\citep{Zhang2025MultiagentAS} further introduces query-specific multi-agent composition using a supernet-like architecture. 
Despite these advances, current frameworks still face two major limitations. First, most multi-agent frameworks remain LLM-homogeneous, relying on a single backbone model (\textit{e.g.}, GPT-4o-mini) for all agents and thus missing the benefits of heterogeneous collaboration across models. Second, they lack complexity diversity: most systems adopt uniformly complex workflows optimized for accuracy, ignoring that real-world queries vary widely in difficulty.

\paragraph{Difficulty-Aware Reasoning.}  
 Recent advances in the reasoning domain of large language models increasingly emphasize difficulty-aware mechanisms to address the limitations of uniform reward signals across heterogeneous tasks, particularly in mathematical reasoning where problem complexity varies widely~\cite{Li2025KnowWT, Zhang2025GRPOLEADAD, Ji2025HowDS, Tong2024DARTMathDR}. These methods dynamically adjust learning objectives based on task difficulty estimates, prioritizing deeper exploration for challenging problems while promoting efficiency on simpler ones. However, the model itself lacks intrinsic knowledge of the difficulty of multi-agent workflows, making it unable to align the difficulty of queries with that of multi-agent workflows. To address this, we propose incorporating difficulty awareness into automatic workflow generation, resolving the perception of workflow difficulty for queries and enhancing workflow adaptability.


\section{Methodology}

\paragraph{Overview.}
Figure~\ref{fig:framework} illustrates our Difficulty-Aware Agentic Orchestration (DAAO), which generates query-specific agentic workflows across domains and difficulty levels. 
Our key contribution is a standalone query difficulty estimator $N_{\theta_d}$ that provides an explicit, calibrated difficulty signal for each input. 
Unlike prior controller networks~\cite{zoph2017nas} that score architectures without reliably assessing the query’s difficulty, $N_{\theta_d}$—instantiated as a variational autoencoder (VAE)~\cite{kingma2014autoencoding} with a learned difficulty head—encodes the query into a latent representation $z$ and outputs a scalar difficulty $d\in(0,1)$. 
This difficulty estimate conditions (i) a layered \emph{operator allocator} $N_{\theta_o}$ that selects an appropriate subset of agentic operators and workflow depth, and (ii) a \emph{cost-aware LLM router} $N_{\theta_m}$ that assigns backbone models by balancing reasoning needs with computational budget. 
The three modules together yield a customized multi-stage workflow per query. 
After execution, we evaluate the output quality and use the success signal to update $N_{\theta_d}$ and refine $N_{\theta_o}$/ $N_{\theta_m}$, enabling continual improvement while keeping difficulty estimation central to workflow construction.

\subsection{Preliminary}
\noindent
This section formalizes the search space for difficulty-aware agentic workflow generation and the cost--utility objective optimized by our policy.

\paragraph{Agentic operator and workflow}
Let $\mathbb{M}$ be the set of available large language models (LLMs) and $\mathbb{S}$ the set of collaboration protocols (e.g., Chain of Thought, Debate, Ensemble). The catalog of feasible operators is the subset $\mathbb{O}\subseteq \mathbb{M}\times \mathbb{S}$. An agentic operator is a pair of one model and one protocol:
\begin{equation}\label{eq:operator}
  O \;=\; \{M,S\},
  \qquad
  M\in \mathbb{M},\ \ S\in \mathbb{S},
  \ \ O\in \mathbb{O}.
\end{equation}

For example, $\{\text{Qwen2-72B},\text{Chain-of-Thought}\}$ denotes step-by-step reasoning on Qwen2-70B, whereas
$\{\text{GPT-4o-mini},\text{Debate}\}$ configures turn-based multi-agent debate.

An agentic workflow can be described as a Directed Acyclic Graph (DAG):
\begin{equation}\label{eq:dag}
  G = \{\mathcal{V}, \mathcal{E}\},
  \quad
  \mathcal{V}\subseteq\mathbb{O},\ \ 
  \mathcal{E}\subseteq\mathcal{V}\times\mathcal{V},
\end{equation}

\noindent
where $\mathcal{V}$ collects instantiated operators (nodes) and $\mathcal{E}$ encodes directed dependencies (edges).
We endow each workflow with a topological layering $\mathcal{V}=\bigcup_{l=1}^{L}\mathcal{V}_{l}$ and restrict edges to go from earlier to later layers; i.e., if $u\in\mathcal{V}_{l}$ and $v\in\mathcal{V}_{l'}$, then $l'>l$.

\paragraph{Layered Policy}
We define the layered selection policy of \textbf{DAAO} over the operator library $\mathbb{O}$ as

\begin{equation}
  \mathcal{A} \;=\; \big\{\{\pi_{l}(O)\}_{O\in\mathbb{O}}\big\}_{l=1}^{L},\qquad
  \pi_{l}(O) \;=\; \mathbb{P}\!\big(O \,\big|\, \mathcal{Q}, z\big),\ \ O\in\mathbb{O},
  \label{eq:def-policy}
\end{equation}

\noindent where $z$ is the latent embedding produced by our difficulty estimator encoded from the input query $\mathcal{Q}$ and adaptively adjusted across layers \(l=1{:}L\). In other words,  \(z=f_{\phi}(\mathcal{Q}, \mathcal{H}_{1:l-1})\) is a learned state summary that encodes the query and the preceding layer-wise history
\(\mathcal{H}_{1:l-1}=\{\mathcal{A}_k\}_{k=1}^{l-1}\) of active operator sets
(\(\mathcal{A}_k \subseteq \mathbb{O}\)).
The policy induces a joint distribution over multi-layer operator configurations:

\begin{equation}
  \mathbb{P}(G \mid \mathcal{Q})
  \;=\;
  \prod_{l=1}^{L}
  \prod_{O\in\mathbb{O}}
  \big(\pi_{l}(O \mid \mathcal{Q}, z)\big)^{\mathbb{I}[\,O\in \mathcal{A}_l\,]},
  \label{eq:joint}
\end{equation}
where \(G=\{\mathcal{V},\mathcal{E}\}\) is the DAG in Eq.~\eqref{eq:dag}, and the node set
is the union of layerwise active sets, respecting the topological layering.

\paragraph{Optimization Objective.}
Given a benchmark dataset $\mathcal{D}$ containing queries $\mathcal{Q}$ and their oracle answers $a$, the objective of \textbf{DAAO} is to learn a query-conditioned policy that balances task utility and inference cost:
\begin{equation}\label{eq:objective}
\underset{\mathbb{P}(G \mid \mathcal{Q})}{\max}\ 
\mathbb{E}_{\substack{(\mathcal{Q}, a) \sim \mathcal{D} \\[2pt] G \sim \mathbb{P}(G \mid \mathcal{Q})}}
\Big[\, U(G; \mathcal{Q}, a) \;-\; \lambda\, C(G; \mathcal{Q}) \,\Big],
\quad \text{s.t.}\ \lambda \ge 0,
\end{equation}

\noindent where $\mathbb{P}(G | \mathcal{Q})$ is a distribution over query-specific workflows, $U(\cdot)$ and $C(\cdot)$ denote the utility (e.g., accuracy) and cost (e.g., token usage, latency) of executing workflow $G$ on query $\mathcal{Q}$, respectively, and $\lambda$ is a trade-off coefficient that balances performance and cost. The outer expectation is taken over the (empirical) data distribution of queries and answers,
while the inner expectation marginalizes the stochastic workflow \(G\sim P(\cdot\mid Q)\).

\subsection{Difficulty-Aware Agent Orchestration}
Given a query $\mathcal{Q}$, our DAAO builds a query-specific workflow by three difficulty-conditioned decisions produced by
$N_{\theta_L}$ (workflow depths), $N_{\theta_o}$ (operator allocation), and $N_{\theta_m}$ (LLM selection).

\paragraph{Difficulty-conditioned decisions.}
Each decision is a probability distribution whose logits are instantiated by the corresponding module and conditioned on the latent difficulty embedding $z$:
\begin{equation}\label{eq:decisions}
  \underbrace{\pi^{(L)}(L \mid \mathcal{Q}, z)}_{\text{workflow depth}(N_{\theta_L})},\quad
  \underbrace{\pi^{(O)}_{l}(O \mid \mathcal{Q}, l, z)}_{\text{operator allocation}(N_{\theta_o})},\quad
  \underbrace{\pi^{(M)}(M \mid \mathcal{Q}, O, z)}_{\text{model selection}(N_{\theta_m})}.
\end{equation}
Here $z$ is produced by our proposed difficulty estimator $N_{\theta_d}$, and a calibrated scalar $d$ is decoded from $z$ (more details in \ref{sec:diffcult}.) In practice, $z$ parameterizes the logits of $N_{\theta_L}$/$N_{\theta_o}$/$N_{\theta_m}$, while $d$ serves as a scalar hardness prior used for thresholding and capacity scaling. Higher $d$ encourages larger-capacity workflows (e.g., more layers or activating more operators), whereas lower $d$ promotes conservative workflows.

\subsection{Query Difficulty Estimator}
\label{sec:diffcult}

To make workflow generation difficulty-aware and balance performance vs. cost per query, we propose a difficulty estimator that guides the subsequent modules. The difficulty estimator $N_{\theta_d}$ maps the input query $\mathcal{Q}$ to a $k$-dimensional latent difficulty representation $z\in\mathbb{R}^k$ by using a variational autoencoder. To encode a given query $\mathcal{Q}$, a lightweight embedding layer $E_{\phi}$ produces a query embedding:

\begin{equation}
    x = E_\phi(\mathcal{Q}) \in \mathbb{R}^{h},
\end{equation}

\noindent where $h$ is the embedding dimension (e.g., $h{=}384$). We then model a Gaussian posterior for the latent difficulty with diagonal covariance, which captures per-dimension uncertainty while remaining stable and efficient to train:
\begin{align}
  \mu(x) = W_{\mu}x + b_{\mu} \in \mathbb{R}^{k},\qquad
  \log \sigma^{2}(x) = W_{\sigma}x + b_{\sigma} \in \mathbb{R}^{k}, \label{eq:enc-affine}
\end{align}
with layer weights $W_{\mu},W_{\sigma}\in\mathbb{R}^{k\times h}$ and bias $b_{\mu},b_{\sigma}\in\mathbb{R}^{k}$, yielding the variational posterior (approximate posterior) over the latent difficulty $z$ given the query embedding $x$:
\begin{equation}
    q(z\mid x)=\mathcal{N}(\mu(x),\,\mathrm{diag}(\sigma^{2}(x))), 
\end{equation}
We sample \(z\) using the reparameterization: 
\begin{equation}
  z \;=\; \mu(x) + \sigma(x)\odot \varepsilon,
  \qquad \varepsilon \sim \mathcal{N}(0, I_{k}),
\end{equation}

\noindent where $\odot$ denotes elementwise multiplication. All stochastic nodes are reparameterized; gradients are taken w.r.t. $\mu$, $\sigma$ via the pathwise estimator. The latent $z\in\mathbb{R}^{k}$ provides a rich difficulty embedding for downstream policies (depth, operator allocation, model selection), while we also decode an interpretable scalar difficulty $d\in(0,1)$ as task difficulty. Concretely, a one-hidden-layer MLP maps $z$ to $d$:

\begin{equation}
d = \operatorname{sigmoid}(W_o^\top \,\mathrm{ReLU}(W_h z + b_h) + b_o) \in (0,1).
\end{equation}

\noindent where $W_h \in \mathbb{R}^{m\times k}$, $b_h \in \mathbb{R}^{m\times 1}$, $W_o \in \mathbb{R}^{1\times m}$ and $b_o \in \mathbb{R}$ are the weights and biases of the hidden and output layers, respectively.

We train our difficulty estimator $N_{\theta_d}$ with a difficulty-guided objective that aligns the decoded difficulty $d$ with the observed outcome $y\in\{0,1\}$ (solved $y=1$, not solved $y=0$), while regularizing the latent embedding $z$:
\begin{equation}\label{eq:loss}
  \mathcal{L}_{\text{diff}}
  =
  \mathcal{L}_{\text{cal}}(d,y) +
  \lambda\, D_{\mathrm{KL}}\!\big(q(z|x)\|p(z)\big),
  \quad
  p(z)=\mathcal{N}(0,I),
\end{equation}

\noindent where, the KL term keeps the approximate posterior $q(z|x)$ close to the standard normal prior $p(z)$, stabilizing the latent space. We adopt a binary cross-entropy (BCE) as the difficulty-calibration term. Since a higher $d$ denotes a \emph{harder} query (thus lower success probability), we calibrate the predicted success probability as $\hat{p}_{\text{succ}}=1-d$ and set:
\begin{equation}
\mathcal{L}_{\text{cal}}(d,y)=\mathrm{BCE}\big(\hat{p}_{\text{succ}},\,y\big)
=-y\log(1-d)-(1-y)\log d.
\end{equation}

\noindent The coefficient $\lambda>0$ balances calibration and regularization.

\subsection{Agentic Operator Allocator.}

Given a query $\mathcal{Q}$, this module constructs a directed acyclic workflow $G = \{\mathcal{V}, \mathcal{E}\}$ from a candidate operator pool $\mathbb{O}\subseteq \mathbb{M}\times \mathbb{S}$, targeting difficulty-aware orchestration that balances performance and cost.

\paragraph{Depth adaptation.}
Let $L_{max}$ denote the maximum depth. We set
\begin{equation}\label{eq:layer}
  L=\max\{1,\,\lceil d\cdot L_{max}\rceil\},
\end{equation}
so that easier queries yield shallower graphs while harder ones trigger deeper chains. 

\paragraph{Layer-wise MoE selection.}
We select operators layer by layer using a lightweight MoE gate and factorize the joint selection over layers as an autoregressive process:
\begin{equation}
  \mathcal{N}_{\theta_o}\!\big(G \mid \mathcal{Q}, z, O\big)
  =
  \prod_{l=1}^{L}\,
  \pi_{l}^{(O)}\!\big(V_{l}\,\big|\,\mathcal{Q}, z, V_{<l}\big), 
  \quad V_{<l}:=\{V_1,\dots,V_{l-1}\}.
\end{equation}
\noindent Here, $V_{<l}:=\{V_1,\dots, V_{l-1}\}$ denotes the history (all previously chosen operator sets), and $\pi_l(\cdot)$ is the layer-$l$ Mixture-of-Experts (MoE) policy \cite{shazeer2017outrageously,huang2024harder} that outputs a subset $V_l$. This factorization means ``choose the current layer conditioned on all previous choices $V_{<l}$, enabling difficulty-aware $z$, query context-dependent $Q$ routing."

\paragraph{Scoring-to-selection with adaptive width.}
We instantiate $\pi_{l}^{(O)}$ so that each layer’s width (number of operators) adapts to the available evidence. First, we translate the query difficulty context and the history into scalar compatibilities for all candidates:

\begin{equation}\label{eq:score}
  S_i \;=\; \operatorname{FFN}\!\Big(
    z \,\Vert\, v(\mathcal{Q}) \,\Vert\, 
    \sum_{O\in V_1}\!v(O) \,\Vert\, \cdots \,\Vert\, \sum_{O\in V_{l-1}}\!v(O)
  \Big),
  \quad O_i\in\mathbb{O},
\end{equation}

\noindent where $v(\cdot)$ is a lightweight embedding (e.g MiniLM~\citep{wang2020minilm} or Sentence-BERT~\citep{reimers2019sentence}) and $\Vert$ denotes concatenation. Larger $S_i$ means $i$-th operator $O_i$ is more compatible with the current layer, given the accumulated context.

We then convert scores into a subset using a cumulative-threshold decoder. Let \(\mathbb{S}=[S_1,\dots,S_{|\mathbb{O}|}]\), and $S_{(1)} \ge S_{(2)} \ge \cdots \ge S_{(|\mathbb{O}|)}$ be the scores in descending order. Using a preset threshold $\tau$, we determine the number of operators as:
\begin{equation}\label{eq:threshold}
    t \;=\; \min\Big\{\,r \in \{1,\dots,|\mathbb{O}|\}\;:\;\sum_{i=1}^{r} S_{(i)} \;>\; \tau\,\Big\}.
\end{equation}

\noindent where $r$ is the \emph{prefix length} (the number of top-ranked operators considered). Operators are activated in descending score order ($S_{(1)}\ge S_{(2)}\ge\cdots$) and the activation proceeds sequentially until the cumulative evidence exceeds $\tau$.
This rule yields an adaptive layer width—higher aggregate confidence activates more operators; lower confidence activates fewer—while enforcing a budget-like constraint via $\tau$.
Together, \eqref{eq:score}–\eqref{eq:threshold} provide a concrete instantiation of the per-layer policy.

\subsection{LLM Router.}

Inspired by~\citep{Ye2025XMASTB,barandoni2024automatingcustomerneedsanalysis}, we leverage model heterogeneity rather than enforcing a single-LLM workflow. After operator selection, each chosen operator \(O_{(i)}\) (for \(i=1,\dots,t\)) is paired with an LLM from a candidate set. We model the per-operator routing as

\begin{equation}\label{eq:llm-routing-factor}
  \mathbb{N}_{\theta_m}\!\big(\{M_{(i)}\}_{i=1}^{t} \,\big|\, \mathcal{Q}, z, \{O_{(i)}\}_{i=1}^{t}\big)
  \;=\;
  \prod_{i=1}^{t} \pi^{(M)}\!\big(M_{(i)} \,\big|\, \mathcal{Q}, z, O_{(i)}\big),
\end{equation}
where \(\pi^{(M)}(\cdot\,|\,\mathcal{Q},z,O_{(i)})\) is the layer-agnostic LLM policy for the \(i\)-th selected operator. For each selected operator, we define a temperature-scaled softmax over LLM candidates indexed by $m\in\{1,\dots,N_M\}$:
\begin{equation}\label{eq:llm-softmax}
  \pi^{(M)}\!\big(M_{(i)}=M_m \,\big|\, \mathcal{Q}, z, O_{(i)}\big)
  \;=\;
  \frac{\exp\!\left(\langle \hat h_{(i)},\, \hat e_m \rangle / T\right)}
       {\sum_{u=1}^{N_M} \exp\!\left(\langle \hat h_{(i)},\, \hat e_u \rangle / T\right)} \, .
\end{equation}

\noindent where $h_{(i)} \;=\; \mathrm{FFN}_{\mathrm{comb}}\!\big(\mathrm{FFN}_q(\mathcal{Q}) \,\Vert\, W_z z \,\Vert\, \mathrm{FFN}_o(O_{(i)})\big)\in\mathbb{R}^d$, is the combined contextual embedding of the query, difficulty, and operator. $e_m \;=\; \mathrm{FFN}_m(M_m)\in\mathbb{R}^d$ is the projected embedding of candidate LLM $M_m$ and $T$ is the temperature parameter controlling the sharpness of the distribution. The dot product \( \langle \cdot, \cdot \rangle \) measures cosine similarity after the embeddings are normalized.

This routeing policy enables the system to route operators to diverse LLMs based on query difficulty and operator context, promoting specialized and adaptive reasoning across the workflow.

\begin{table*}[!ht]
\caption{
Performance comparison across baseline prompting strategies, single-agent methods, autonomous agentic workflows, and LLM routing approaches. Bold numbers indicate the best performance, while underlined numbers denote the second-best. The LLM pool comprises both lightweight and high-capacity models to support diverse routing strategies.
}
\label{tab:rq1_performance}
\begin{tabular}{l|c|cccccc}
\toprule
{\textbf{Method}}&\textbf{LLM} & \textbf{MMLU} & \textbf{GSM8K} & \textbf{MATH} & \textbf{HumanEval} & \textbf{MBPP} & {\textbf{Avg.}} \\
\midrule
& gpt-4o-mini & 77.81 & 87.45 & 46.29 & 85.71 & 72.20 & 73.89 \\
&  qwen-2-72b & 80.22 & 85.40 & 46.10 & 64.65 & 73.90 & 70.05 \\
& gemini-1.5-flash & 80.04 & 86.76 & 48.00 & 82.61 & 73.00 & 74.08 \\
\multirow{-4}{*}{Vanilla} &  llama-3.1-70b & 79.08 & 86.68 & 45.37 & 80.75 & 68.20 & 72.01 \\
\midrule

 & gpt-4o-mini & 78.43 & 87.10 & 46.40 & 86.69 & 69.60 & 73.64 \\
\multirow{-2}{*}{ CoT~\cite{cot}} & gemini-1.5-flash & 81.35 & 86.47 & 48.00 & 81.37 & 73.00 & 74.04\\

& gpt-4o-mini & 81.05 & 86.89 & 46.53 & 87.58 & 75.80 & 75.57 \\
\multirow{-2}{*}{ComplexCoT~\cite{fu2022complexity}} &  gemini-1.5-flash & 80.74 & 86.01 & 48.28 & 80.12 & 71.80 & 73.39\\

 & gpt-4o-mini & 81.05 & 87.57 & 47.91 & 87.58 & 73.00 & 75.42 \\
\multirow{-2}{*}{SC(CoT)~\cite{wang2023selfconsistency}} & gemini-1.5-flash & 81.66 & 87.50 & 48.73 & 80.75 & 72.00 & 74.13 \\
\midrule

& gpt-4o-mini & 79.54 & 86.12 & 43.18 & 84.19 & 68.13 & 72.23 \\
\multirow{-2}{*}{ADAS~\cite{hu2024adas}} &  gemini-1.5-flash & 79.68 & 86.00 & 45.89 & 80.69 & 68.00 & 72.05\\

 & gpt-4o-mini & 83.10 & 91.16 & 51.82 & 90.93 & 81.67 & 79.73 \\
\multirow{-2}{*}{AFlow~\cite{zhang2024aflow}} & gemini-1.5-flash & 82.35 & 90.43 & 52.00 & 85.69 & 76.00 & 77.29 \\

& gpt-4o-mini & 83.01 & \underline{92.30} & 51.82 & \underline{92.85} & 82.17 & 80.43 \\
\multirow{-2}{*}{MaAS~\citep{Zhang2025MultiagentAS}} &  gemini-1.5-flash & 83.42 & 92.00 & 52.25 & 90.55 & 82.69 & 80.18\\

\midrule
PromptLLM~\citep{feng2024graphroutergraphbasedrouterllm} & LLM Pool & 78.43 & 88.68 & 52.30 & 86.33 & 73.60 & 75.86\\ 

RouteLLM~\citep{ong2024routellmlearningroutellms} & LLM Pool & 81.04 & 89.00 & 51.00 & 83.85 & 72.60 & 75.50\\

MasRouter~\citep{Yue2025MasRouterLT} & LLM Pool & \underline{84.25} & 92.00 & \underline{52.42} & 90.62 & \underline{84.00} & \underline{80.66}\\ 
\midrule

\textbf{Ours} & LLM Pool & \textbf{84.90} & \textbf{94.40} & \textbf{55.37} & \textbf{94.65} & \textbf{86.95} & \textbf{83.26}\\
\hline
\bottomrule
\end{tabular}
\end{table*}

\section{Experiments}

\subsection{Experiment Setup}
\paragraph{Benchmarks.} We evaluate \textbf{DAAO} on six public benchmarks covering three domains: (1) math reasoning, GSM8K~\citep{gsm8k} and MATH~\citep{hendrycksmath2021}; (2) code generation, HumanEval~\citep{human-eval} and MBPP~\citep{austin2021mbpp}); GAIA~\citep{mialon2023gaiabenchmark}. Additionally, we include MMLU~\citep{mmlu}, a benchmark covering 57 academic subjects, to assess general knowledge and multitask language understanding. For the MATH benchmark, we follow \cite{hong2024datainterpreter} in selecting a harder subset (617 problems). The dataset metric are in Appendix~\ref{app:metric}.

\paragraph{Baselines.} We compare \textbf{DAAO} with three of agentic baselines: (1) single-agent approaches, including CoT~\citep{cot}, ComplexCoT~\citep{fu2022complexity}, Self-Consistency~\citep{wang2023selfconsistency}; (2) autonomous agentic workflows, including ADAS~\citep{hu2024adas}, AFlow~\citep{zhang2024aflow} and MaAS~\citep{Zhang2025MultiagentAS}. (3) LLM routers, PromptLLM~\citep{feng2024graphroutergraphbasedrouterllm}, RouteLLM~\citep{ong2024routellmlearningroutellms} and MasRouter~\citep{Yue2025MasRouterLT}.

\paragraph{LLM Backbones.} We select LLM Pool with varying sizes and capacities, including gpt-4o-mini-0718~\cite{OpenAI-gpt4o}, gemini-1.5-flash~\citep{geminiteam2024gemini15unlockingmultimodal}, llama-3.1-70b~\citep{dubey2024llama}, Qwen-2-72b~\cite{Yang2024Qwen2TR}. LLMs are accessed via APIs, with the temperature set to 1. We selected gpt-4o-mini-0718~\cite{OpenAI-gpt4o} and gemini-1.5-flash~\citep{geminiteam2024gemini15unlockingmultimodal}, which performed well in Vanilla, as the models for other baselines.

\paragraph{Implementation Details.} Building upon established methodologies in workflow automation~\cite{saad2024archon,hu2024adas,zhang2024aflow}, we divide each dataset into training and test sets using a \textsc{train:test} ratio of 1:4. We initialize the feasible space of operator nodes with the following operators: CoT, LLM-Debate, Review, Ensemble, ReAct, Self-Consistency, Testing. Detailed instructions are in Appendix~\ref{app:operator}. We set the max number of layers as $L_{max}$ = 5, the cost penalty coefficient $\lambda$ as $\lambda\in\{1e-3,5e-3,1e-2\}$, the sampling times $K$ = 4 and threshold \( \tau \) = 0.3. To ensure robust results, we conducted each experiment three times and reported the average performance.

\subsection{Performance Analysis}

\subsubsection{High-performing.}
The experimental results in Table~\ref{tab:rq1_performance} demonstrate that DAAO effectively constructs high-performing agentic workflows. Compared to existing automated orchestration methods, DAAO achieves an average accuracy improvement of $3.5\% \sim 15.2\%$, and outperforms recent LLM routing methods by $3.2\% \sim 10.2\%$. On the MATH benchmark, DAAO attains a best-in-class score of 55.37\%, surpassing the second-best method, MasRouter, by 2.95\%. Across five datasets, DAAO consistently outperforms all baselines, highlighting its versatility and robustness.

Table~\ref{tab:rq1_gaia} further compares DAAO with existing automated systems on the GAIA benchmark—a challenging, high-complexity evaluation suite for multi-agent systems in realistic, multimodal, and tool-augmented settings. Unlike traditional benchmarks focused on static question answering or single-step reasoning, GAIA tasks require multi-step planning, cross-modal understanding, and tool interaction (e.g., web browsing, file system access). While AFlow uses a fixed workflow and MaAS does not fully exploit LLM specialization, DAAO dynamically generates query-specific workflows and allocates tasks to LLMs based on domain expertise. As a result, DAAO outperforms AFlow and MaAS by 17.97\% and 8.33\%, respectively, demonstrating its effectiveness in complex, real-world scenarios.

\begin{table}[!ht]
\caption{Performance comparison on the GAIA benchmark. Results are reported across three difficulty levels, with average scores shown in the last column. The best results are highlighted in bold.}
\label{tab:rq1_gaia}
\begin{tabular}{l|cccc} 
    \toprule
    \textbf{Method} & \textbf{Level 1} & \textbf{Level 2} & \textbf{Level 3}& \textbf{Avg.} \\ \toprule
    GPT-4o-mini  & $7.53$ & $4.40$ & $0$ & $4.65$\\
    \midrule
    ADAS & $13.98$ & $4.40$ & $0$ & $6.69$ \\
    AFlow & $10.75$ & $8.81$ & $4.08$ & $8.00$  \\
    MaAS & $20.45$ & $18.61$ & $6.25$ & $17.64$  \\
    \midrule
    \textbf{Ours}  & \textbf{$\mathbf{30.42}$} & \textbf{$\mathbf{24.00}$} & $\mathbf{8.50}$ &  $\mathbf{25.97}$ \\
    \bottomrule
\end{tabular}
\end{table}

\begin{table}[!ht]
\caption{Training, inference, and overall cost (in USD) on the MATH benchmark, along with corresponding accuracy. Our method achieves the lowest cost and highest accuracy. AFlow and MaAS use GPT-4o-mini, while other methods utilize an LLM pool.}

\label{tab:cost}
\begin{tabular}{l|cccc} 
    \toprule
    \textbf{Method} & \textbf{Training} & \textbf{inference} & \textbf{overall}& \textbf{Acc.} \\ \toprule
    AFlow & $22.50$ & $1.66$ & $24.16$ & $51.82$  \\
    MaAS & $3.38$ & $0.42$ & $3.80$ & $51.82$  \\
    MasRouter & $3.56$ & $0.65$ & $4.21$ & $52.42$  \\
    \midrule
    \textbf{Ours}  & \textbf{$\mathbf{2.34}$} & \textbf{$\mathbf{0.27}$} & $\mathbf{2.61}$ &  $\mathbf{55.37}$ \\
    \bottomrule
\end{tabular}
\end{table}

\subsubsection{Cost-effective.}
We emphasize the cost-efficiency of our agentic automation framework across two key dimensions: training expenditure and inference overhead. We compare against AFlow and MaAS, where AFlow represents the state-of-the-art (SOTA) among task-level frameworks, and MaAS is the SOTA among query-level frameworks. As shown in Table~\ref{tab:cost}, AFlow incurs a substantial training cost of \$22.50 and inference cost of \$1.66, totaling \$24.16. In contrast, our method significantly reduces these costs to \$2.34 for training and \$0.27 for inference—only 10.4\% and 16.3\% of AFlow's respective costs. MasRouter adopts a collaborative paradigm, assigning multiple LLMs to role-play in handling a query. However, it suffers from two drawbacks: (1) redundant participation of LLMs in each collaborative step, and (2) lack of adaptation for easy queries, leading to excessive cost without proportional performance gains.
Notably, our method not only reduces cost but also achieves the highest accuracy of 55.37\%, outperforming both AFlow and MaAS.


This cost-efficiency is attributed to two strategies: (1) Our difficulty-awareness strategy, which generates adaptive workflows for queries, employing simple workflows for easy queries while allocating more resources to the generation of workflows for more complex queries; (2) Our adaptive model selection strategy, which dynamically leverages more affordable models (such as LLaMA-3.1 or Qwen-2-72B) when sufficient, rather than defaulting to high-cost models like GPT-4o-mini.

\begin{figure*}[!ht]
\centering
\includegraphics[width=1\linewidth]{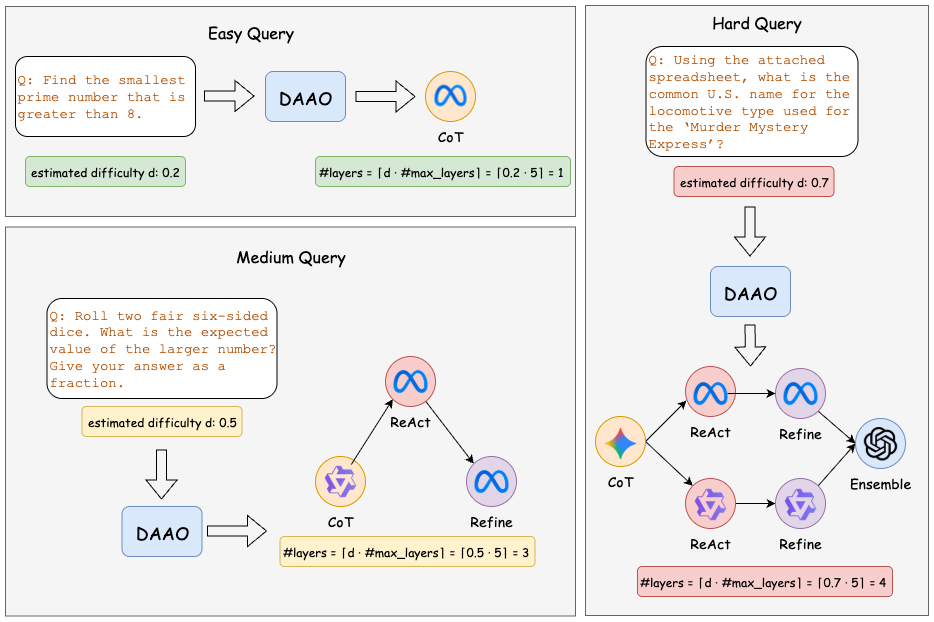}
\caption{The visualization of the workflow generated by DAAO. Colors represent different models assigned to each operator.
}
\label{fig:case}
\end{figure*}

\subsection{Case Study}
As shown in Figure \ref{fig:case}, DAAO generates different workflows for queries of varying difficulty. For simple queries, it produces streamlined workflows, sometimes using only a single operator. For medium or difficult queries, it constructs deeper and more complex workflows by exploring a broader combination of operators. This demonstrates DAAO’s difficulty-aware paradigm: selecting economical workflows for simple queries to enable rapid completion, while leveraging sophisticated workflows for complex queries to meet higher demands.

\begin{table}[h]
    \centering
    \caption{Cross-domain optimization performance}
    \label{tab:cross-domain}
    \begin{tabular}{ccr}
        \toprule
        \textbf{Train on} & \textbf{Test on} & \textbf{Perf.} \\
        \midrule
        MATH & MATH & 55.37 \\
        MATH & GSM8K & 95.44 \\
        \midrule
        MATH+GSM8K & MATH & 56.42 \\
        MATH+GSM8K & GSM8K & 95.70 \\
        \midrule
        HumanEval & HumanEval & 94.65 \\
        HumanEval & MATH & 54.46 \\
        \midrule
        HumanEval+MATH & HumanEval & 95.00 \\
        HumanEval+MATH & MATH & 55.50 \\
        \bottomrule
    \end{tabular}
\end{table}

\subsection{Inductive Ability Analysis}
\subsubsection{Cross-domain Optimization.}
We present the performance results of our cross-domain training experiments (see Table \ref{tab:cross-domain}), evaluating the impact of multi-domain joint optimization on generalization. We observe that single-domain training achieves baseline performance on the MATH dataset, with strong transfer to GSM8K, primarily due to their shared mathematical reasoning requirements. Moreover, since MATH problems are generally more challenging than those in GSM8K, training on MATH equips the framework with advanced problem-solving skills that effectively generalize to the relatively simpler GSM8K. In contrast, we find that single-domain training on HumanEval yields high fidelity on code generation (94.65\%) but limited transfer to MATH (54.46\%), underscoring domain-specific overfitting. Notably, we achieve modest improvements across target tasks (approximately 0.35\%–1.05\%) through joint training (e.g., MATH+GSM8K and HumanEval+MATH), without inducing catastrophic forgetting, indicating that our simultaneous exposure to multiple domains promotes shared representation learning and enhances model robustness. 

Overall, our multi-domain setup not only preserves intra-domain proficiency but also slightly boosts cross-domain generalization. Furthermore, within the same domain, training on more complex benchmark datasets can further enhance the framework's ability to generate workflows for complex queries.

\begin{figure}[!ht]
\includegraphics[width=0.95\linewidth]{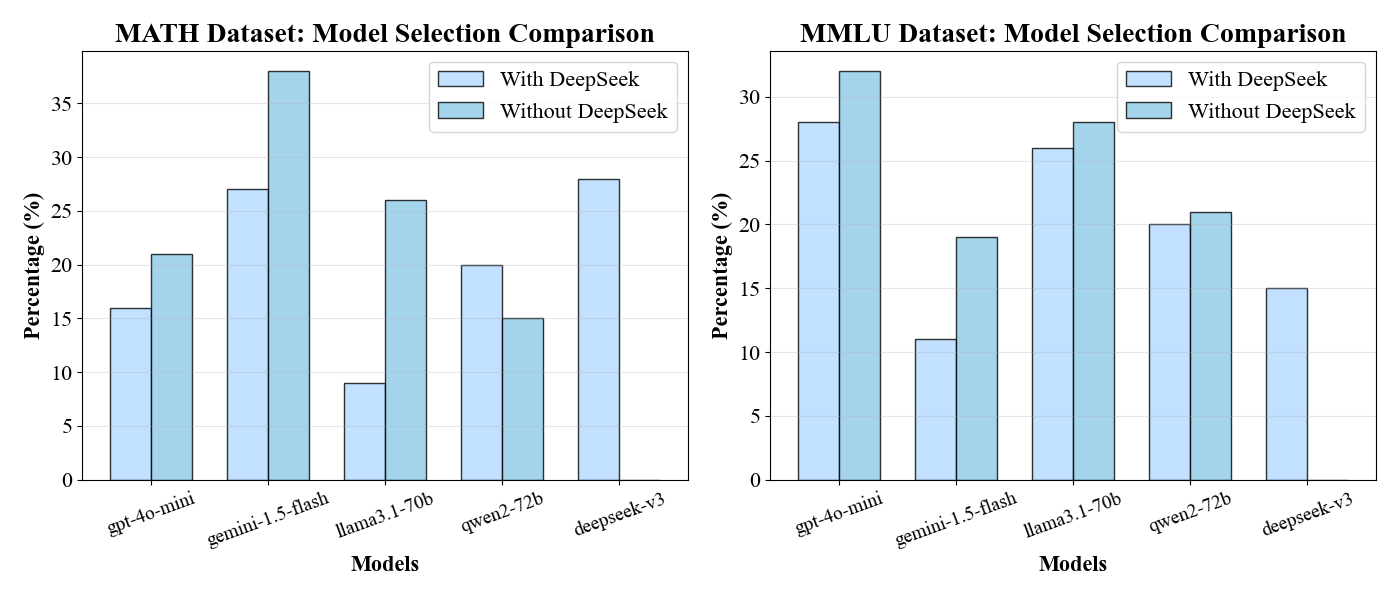}
\caption{Distribution of LLM selections made by \textbf{DAAO} on the MATH and MMLU benchmarks. }
\label{fig:llmselect}
\end{figure}

\subsubsection{LLM Router Analysis.}
In this section, we validate that DAAO does not exhibit a preference for any particular LLM and demonstrate its ability to generalize well to unseen LLMs without requiring extensive pretraining. Figure \ref{fig:llmselect} illustrates the distribution of LLMs selected by DAAO on the MATH and MMLU datasets before and after the addition of DeepSeek-v3, with the new model being chosen 29\% and 15\% of the time, respectively. By intelligently selecting models that match the difficulty and domain of each query, DAAO improved the accuracy on MATH from 55.37\% to 56.20\% and increased the accuracy on MMLU from 84.90\% to 85.66\%.

\begin{table}[!htpb]
\caption{Ablation study of \textbf{DAAO} on HumanEval and MATH. We report performance (Pass@1 or Accuracy) and corresponding inference cost. \textit{w/o} $\textbf{DA}$ removes the difficulty-aware module; \textit{w/o} $\textbf{LS}$ disables LLM selection; \textit{w/o} $\textbf{C}(\cdot)$ omits the cost-awareness component.}
\label{tab:ablation}
\begin{tabular}{c|cc|cc}
\toprule
\makecell{Dataset} &\multicolumn{2}{c|}{HumanEval} & \multicolumn{2}{c}{MATH}\\
\midrule
\makecell{Metric}  &  \makecell{Pass@1 \\(\%)} & \makecell{Cost\\ ($10^{-3}\$$)}   &  \makecell{Accuracy \\(\%)} & \makecell{Cost\\ ($10^{-3}\$$)}  \\
\midrule
\makecell{Vanilla}    & $94.65$ & $1.10$ & $55.37$ & $0.55$ \\
\midrule
 \textit{w/o} $\textbf{DA}$  & $92.21$ & $1.64$ & $52.18$ & $0.88$ \\
\textit{w/o} $\textbf{LS}$ & $92.69$ & $1.38$ &  $53.24$ & $0.79$   \\
 \textit{w/o} $\textbf{C}(\cdot)$ & $94.72$ & $1.88$ & $55.40$ & $1.00$\\
\bottomrule
\end{tabular}
\end{table}
\subsection{Framework Analysis}
\subsubsection{Ablation Study.}
We conduct an ablation study on three key components of our DAAO framework:
(1) w/o \textbf{DA}, removing the difficulty-aware module; and (2) w/o \textbf{LS}, removing the LLM selector and routing all subtasks to a fixed LLM;(3) w/o $\textbf{C}(\cdot)$, eliminating the cost constraint in~\Cref{eq:objective}.
As shown in Table~\ref{tab:ablation}, removing the difficulty-aware module leads to the largest drop in both accuracy and efficiency, especially on the MATH dataset. This highlights the importance of adaptive reasoning control based on estimated query difficulty (e.g., dynamically adjusting the number of reasoning layers instead of using a fixed number).
Removing the LLM router slightly affects accuracy, but leads to a notable increase in inference cost, as it prevents the system from using lightweight models when appropriate. Removing \(\textbf{C}(\cdot)\) does not significantly impact the performance, but it disrupts the adaptive capability of DAAO to query difficulty.
Overall, the results demonstrate that both components are crucial for balancing performance and cost in multi-step reasoning tasks.

\begin{figure}[!ht]
\includegraphics[width=1\linewidth]{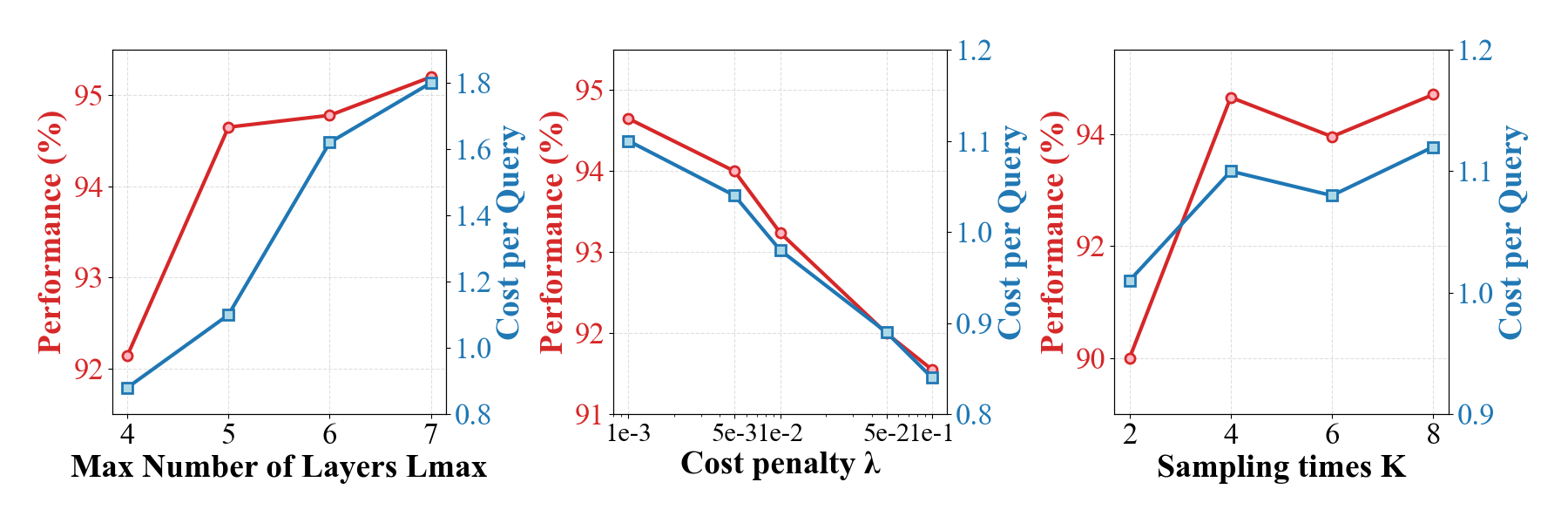}
\caption{Sensitivity analysis of DAAO on HumanEval. The unit of cost per query (right) and performance (left) is $10^{-3}$ · \$ and pass@1 (\%), respectively.}
\label{fig:Sensitivity}
\end{figure}

\begin{table}[t]
  \centering
  \caption{Performance(\%) and Average Cost($10^{-3}\$$) across thresholds \( P \) on HumanEval and GSM8K.}
  \label{tab:threshold}
  \resizebox{\columnwidth}{!}{%
  \begin{tabular}{l l c c c c c c c}
    \toprule
    Dataset & Metric & 0.1 & 0.2 & 0.3 & 0.4 & 0.5 & 0.6 & 0.7 \\
    \midrule
    \multirow{2}{*}{HumanEval} 
      & Perf.     & 92.80 & 93.60 & 94.65 & 94.42 & 94.80 & 94.77 & 94.80 \\
      & Cost & 0.86  & 0.94  & 1.10  & 1.29 & 1.28 & 1.78 & 2.30 \\
    \midrule
    \multirow{2}{*}{GSM8K} 
      & Perf.     & 91.99 & 93.46 & 94.40 & 94.34 & 94.70 & 94.75 & 94.40 \\
      & Cost & 0.40  & 0.45  & 0.50  & 0.59  & 0.68  & 0.77  & 0.96 \\
    \bottomrule
  \end{tabular}%
  }
\end{table}

\subsubsection{Sensitivity Analysis.}
We analyzed the sensitivity of DAAO to four core parameters: the maximum number of layers in the agentic supernet $L_{max}$ in Equation \Cref{eq:layer}, the cost penalty coefficient $\lambda$ in~\Cref{eq:objective}, the sampling count $K$ and the threshold \( \tau \) in Equation \Cref{eq:threshold}. The results are shown in Figure~\ref{fig:Sensitivity} and Table~\ref{tab:threshold}. For the parameter $L_{max}$, we observed a significant performance improvement when $L_{max}$ increased from 4 to 5 (from 92.9\% to 94.6\%). However, further increases in $L_{max}$ only yielded marginal performance gains while significantly increasing the inference cost per query. Considering both performance and cost, we selected $L_{max}$ = 5. For the parameter $\lambda$, we found that larger $\lambda$ values led DAAO to favor more cost-efficient solutions, but with a slight performance degradation. For the parameter $K$, we note that performance is suboptimal with highest variance when $K$ = 2. Increasing $K$ to 4 effectively achieves a satisfactory low-variance estimation. For the threshold \( \tau \), as shown in Table \ref{tab:threshold}, performance improves as the threshold \( \tau \) increases. However, the gain stops growing beyond 0.3. At the same time, a higher threshold \( \tau \) raises inference costs because more operators are activated in each layer. Therefore, we set the threshold \( \tau \) to 0.3 to balance performance and efficiency.

\section{Conclusion}
In this work, we propose DAAO, a difficulty-aware agentic orchestration framework that dynamically adapts reasoning workflows to query complexity and domain. By estimating query difficulty, allocating modular operators, and routing to heterogeneous LLMs, DAAO builds flexible and cost-efficient workflows. It moves beyond one-size-fits-all designs by leveraging complementary LLM strengths and adapting workflow depth per query. Experiments across six benchmarks show that DAAO outperforms existing multi-agent and routing systems, achieving up to 11.21\% higher accuracy with up to 36\% lower cost. These results highlight the value of difficulty-guided modular orchestration for scalable LLM agents. Future work includes extending DAAO to multi-modal queries and incorporating real-time feedback.

\section{Acknowledgement}
This work was supported by the National Natural Science Foundation of China under Grant 52308250 and Autoagents.AI.


\bibliographystyle{ACM-Reference-Format}
\balance
\bibliography{sample-base}

\appendix

\section{Notations}\label{app:notations}

\begin{itemize}
\item \textbf{$O = {M,S}$}: An agentic operator defined by pairing an LLM $M$ with a protocol $S$.
\item \textbf{$M$}: A large language model (LLM) instance.
\item \textbf{$\mathbb{M}$}: The set of all available LLMs.
\item \textbf{$S$}: A collaboration or reasoning protocol (e.g., Chain-of-Thought, Debate).
\item \textbf{$\mathbb{S}$}: The set of all feasible protocols.
\item \textbf{$\mathbb{O} \subseteq \mathbb{M}\times\mathbb{S}$}: The catalog of feasible agentic operators.
\item \textbf{$G = {\mathcal{V}, \mathcal{E}}$}: A directed acyclic graph (DAG) representing an agentic workflow, with operator nodes $\mathcal{V}$ and dependency edges $\mathcal{E}$.
\item \textbf{$\mathcal{V} = \bigcup_{\ell=1}^{L}\mathcal{V}{\ell}$}: The layered decomposition of workflow nodes across $L$ layers.
\item \textbf{$L$}: The number of workflow layers (depth).
\item \textbf{$\mathcal{A} \;=\; \big\{\{\pi_{l}(O)\}_{O\in\mathbb{O}}\big\}_{l=1}^{L}$}: Layered operator-selection policy, where $\pi_{\ell}(O)$ is the probability of selecting operator $O$ at layer $\ell$.
\item \textbf{$\pi^{(L)}(L \mid \mathcal{Q}, z)$}: Distribution over workflow depth given $\mathcal{Q}$ and $z$.
\item \textbf{$\pi^{(O)}_{l}(O \mid \mathcal{Q}, l, z)$}: Distribution over operators at layer $\ell$ conditioned on $\mathcal{Q}$ and difficulty embedding $z$.
\item \textbf{$\pi^{(M)}(M \mid \mathcal{Q}, O, z)$}: Model-routing policy assigning an LLM $M$ to operator $O$ based on $\mathcal{Q}$ and $z$.
\item \textbf{$z \in \mathbb{R}^{k}$}: Latent difficulty embedding of the query produced by the difficulty estimator $N{\theta_d}$.
\item \textbf{$d \in (0,1)$}: Scalar difficulty score decoded from $z$, where larger $d$ indicates a harder query.
\item \textbf{$N_{\theta_d}$}: Difficulty estimator (a VAE with a learned difficulty head).
\item \textbf{$N_{\theta_o}$}: Operator allocator that selects operators per layer.
\item \textbf{$N_{\theta_m}$}: Cost-aware LLM router that selects models for operators.
\item \textbf{$N_{\theta_L}$}: Workflow depth selector determining $L$ based on $z$.
\item \textbf{$U(G; \mathcal{Q}, a)$}: Utility (e.g., accuracy) achieved by executing workflow $G$ on query $\mathcal{Q}$.
\item \textbf{$C(G; \mathcal{Q})$}: Inference cost (e.g., token usage, latency) of workflow $G$ for query $\mathcal{Q}$.
\item \textbf{$\lambda$}: Trade-off coefficient balancing utility and cost in Eq.~\eqref{eq:objective}.
\item \textbf{$v(\cdot)$}: Text or operator embedding function (e.g., MiniLM, SBERT).
\item \textbf{$\tau$}: Cumulative evidence threshold controlling layer width (Eq.~\eqref{eq:threshold}).
\item \textbf{$K$}: The number of samples in each data set.
\end{itemize}

\section{Technical Details}

\subsection{Operator Space}\label{app:operator}
In this section, we detail the initialization of operator nodes as follows:

\begin{enumerate}
    \item \textbf{Chain-of-Thought (CoT).}  
    CoT~\cite{cot} reasoning encourages the LLM to think step by step rather than directly outputting an answer. This approach enhances its capability to solve complex problems through intermediate reasoning steps, improving task handling and providing greater transparency in the decision-making process.
    \item \textbf{LLM-Debate.}  
    LLM-Debate~\cite{arXiv2023_MultiAgent-Debate} allows multiple LLMs to debate, leveraging diverse perspectives to identify better solutions. In practice, we initialize three debaters and permit up to two debate rounds.
    \item \textbf{Self-Consistency.}  
    Adopting the methodology from \cite{wang2023selfconsistency}, this operator aggregates five CoT reasoning paths and determines the final answer through majority voting.
    \item \textbf{Self-Refine.}  
    Following \cite{NeurIPS2023_Self-Refine}, this operator initially generates an answer using CoT reasoning, then prompts the agent to self-reflect iteratively. We set a maximum of five refinement iterations.
    \item \textbf{Ensemble.}  
    Inspired by LLM-Blender~\cite{blender}, this operator involves three LLM-powered agents from different sources outputting answers to the same query. The pairwise ranking is used to evaluate and aggregate their responses into a final solution.
    \item \textbf{Testing.} Following the test designer in AgentCoder~\cite{huang2023agentcoder}, this operator is used for generating test cases for the generated code.
    \item \textbf{ReAct.} Following~\cite{yao2023react}, this operator enables the agent to leverage versatile tools, including code interpreter, web searching, external knowledge database, \textit{etc.}, to handle diverse user demands. 
\end{enumerate}

\subsection{Embedding Function}\label{app:embedding}

Following established practices~\cite{feng2024graphroutergraphbasedrouterllm}, we first employ an LLM to generate a comprehensive profile description for each operator. Subsequently, a lightweight text embedding model (in our case, MiniLM~\citep{wang2020minilm}) is used to encode the profile into a fixed-dimensional embedding. The prompt for generating the operator profile is as follows:

\begin{tcolorbox}[notitle, sharp corners, breakable, colframe=Periwinkle, colback=white, 
       boxrule=3pt, boxsep=0.5pt, 
       title={Embedding Prompt},]\label{box:operator-profile}
       \footnotesize
       {\fontfamily{pcr}\selectfont
\begin{lstlisting}
prompt = """You are a highly proficient expert in designing and defining operators for large language models (LLMs). Your primary objective is to meticulously generate the `description` and `interface` fields for a specified operator based on its provided Python implementation. The generated content must be accurate, efficient, and precisely reflect the functionality of the operator's code.

To ensure consistency, quality, and adherence to best practices, refer to the following examples of previously defined operators:
{
    "Generate": {
        "description": "Generates anything based on customized input and instruction.",
        "interface": "generate(input: str, instruction: str) -> dict with key 'response' of type str" 
    },
    "ScEnsemble": {
        "description": "Uses self-consistency to select the solution that appears most frequently in the solution list, improving the selection to enhance the choice of the best solution.",
        "interface": "sc_ensemble(solutions: List[str], problem: str) -> dict with key 'response' of type str"
    }
}

Now, given the following operator code. This code encompasses the function signature, parameters with type annotations, internal logic, and return statements essential for comprehensively understanding the operator's purpose and behavior.Please provide its `description` and `interface` fields in the same format.
[operator code]

 """
\end{lstlisting}
}
\end{tcolorbox}

\section{Experimental Details}\label{app:metric}
In this section, we introduce each dataset along with its primary evaluation metric.

\begin{itemize}
    \item \textbf{HumanEval}: uses \textbf{Pass@k} as the primary metric to measure the proportion of correctly generated samples in code generation tasks.
    \item \textbf{MBPP}: also uses \textbf{Pass@k} to evaluate the model's ability to generate solutions for programming problems.
    \item \textbf{GSM8K}: uses \textbf{Accuracy} to assess the model's correctness in mathematical reasoning problems.
    \item \textbf{MATH}: uses \textbf{Accuracy} as the core metric to evaluate the model's performance on solving mathematics problems.
    \item \textbf{MMLU}: uses \textbf{Accuracy} to measure the model's performance on multi-domain knowledge question-answering tasks.
    \item \textbf{GAIA}: uses \textbf{Accuracy} as the primary metric to evaluate the model's performance on general AI tasks.
\end{itemize}

\end{document}